\documentclass{article}

\usepackage[preprint]{neurips_2026}

\usepackage[utf8]{inputenc} 
\usepackage[T1]{fontenc}    
\usepackage{url}            
\usepackage{booktabs}       
\usepackage{amsfonts}       
\usepackage{nicefrac}       
\usepackage{microtype}      
\usepackage{xcolor}         
\usepackage{graphicx}
\usepackage{subcaption}
\usepackage{wrapfig}

\usepackage{amsmath}
\usepackage{amssymb}
\usepackage{mathtools}
\usepackage{amsthm}

\definecolor{lightblue}{RGB}{217, 237, 247}
\definecolor{lightgreen}{RGB}{229, 245, 224}

\usepackage{hyperref}
\usepackage[capitalize,noabbrev]{cleveref}

\theoremstyle{plain}
\newtheorem{theorem}{Theorem}[section]

\theoremstyle{definition}

\theoremstyle{remark}

\usepackage{makecell}
\usepackage{multirow}
\usepackage[table]{xcolor} 
\usepackage[textsize=tiny]{todonotes}
\title{Flexformer: Flexible Linear Transformer with Learnable Attention Kernel}

\author{%
  Haoran Zhang, Feng Zhou\thanks{Corresponding author.} \\
  Center for Applied Statistics and School of Statistics, Renmin University of China\\
  Beijing, China \\
  \texttt{zhrrhz2002@163.com}, \texttt{feng.zhou@ruc.edu.cn}
}

\begin{document}

\maketitle

\begin{abstract}
  Transformer models rely on attention mechanism to capture long-range dependencies but suffer from quadratic complexity, limiting their scalability to long sequences. Kernel-based linear attention reduces this complexity but typically relies on fixed or weakly learnable kernels, restricting expressiveness and performance. In this work, we propose Flexformer, a flexible linear Transformer that learns attention kernels in a fully data-driven manner. Flexformer builds on random Fourier feature-based linear attention and treats spectral frequencies as trainable parameters, enabling the model to learn a broad family of attention kernels. 
  We develop both stationary and nonstationary variants, with the latter offering strictly greater expressiveness. 
  Extensive experiments on language modeling and sequence classification demonstrate that Flexformer consistently outperforms baselines. Moreover, Flexformer can be effectively distilled from pretrained Transformers to recover softmax attention and exhibits strong kernel transferability across domains, achieving both high efficiency and competitive performance on long-sequence tasks.
\end{abstract}

\section{Introduction}
The Transformer~\cite{transformer} architecture has achieved remarkable success in machine translation tasks since its original proposal. Subsequently, numerous variants of Transformer have been extensively applied across prominent domains within artificial intelligence, including natural language processing~\cite{BERT,roberta,fewshot}, computer vision~\cite{vit,Swin}, time series analysis~\cite{informer,PatchTST,itransformer}, and audio processing~\cite{wav2vec,CVTransformer}. 
The most critical component in these Transformer-based models is the dot-product attention mechanism. It operates by computing pairwise token affinities across the entire input sequence, thereby enabling each token to dynamically aggregate the most relevant information pertaining to itself, while effectively modeling long-range dependencies within the sequence. 
Unlike sequential RNNs~\cite{RNN} processing tokens step-by-step, Transformer attention computes representations for all positions in parallel, maximizing computational hardware utilization.

This effective attention mechanism, however, suffers from efficiency limitations: it computes attention weights for all pairs of positions in the sequence, leading to both time and space complexity that grow quadratically with the sequence length. 
To address this issue, numerous studies have attempted to optimize the computation of the attention mechanism. These redesigned or optimized attention computation methods should effectively reduce time and space complexity while achieving performance that comparable or even surpasses that of softmax attention. 

\begin{figure*}[!t]
    \centering
    \begin{subfigure}[b]{0.46\textwidth}
        \includegraphics[
            trim=70pt 100pt 70pt 80pt,  
            clip,
            width=\textwidth
        ]{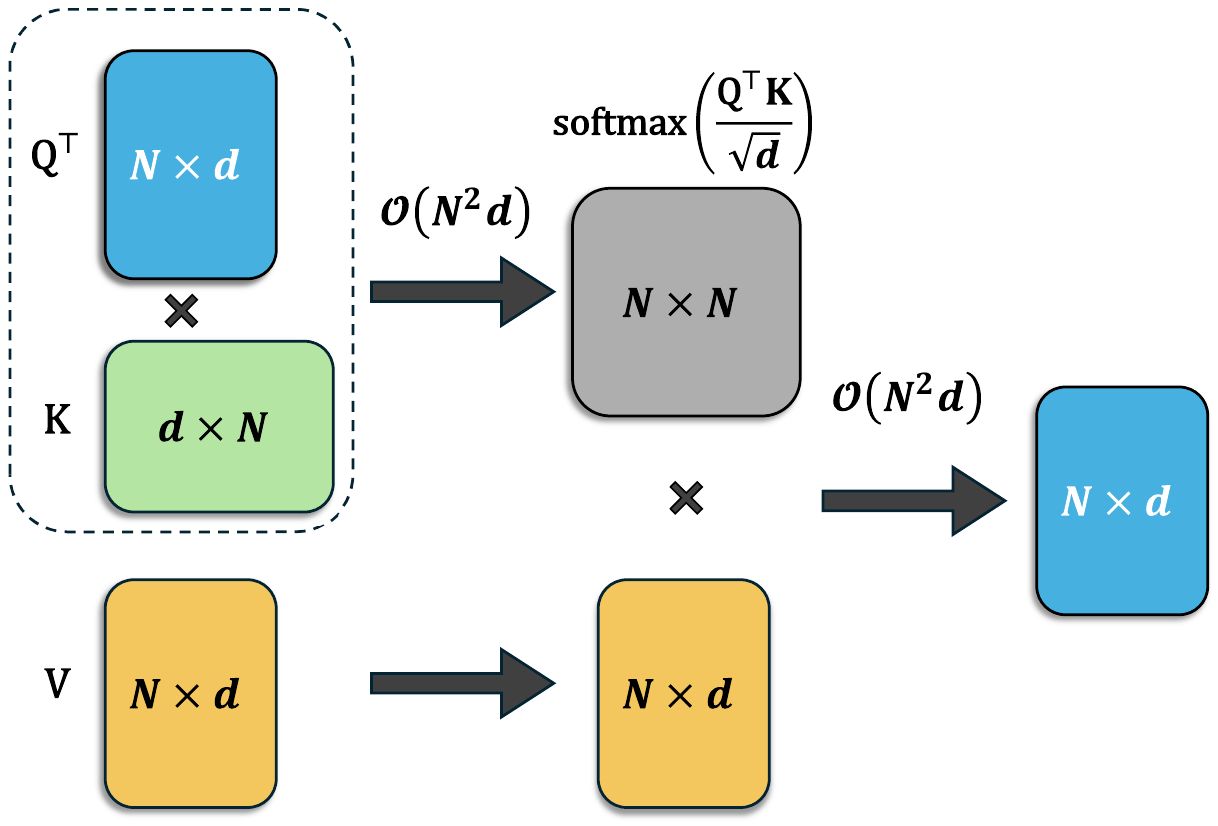}
        \caption{Softmax attention.}
        \label{cg1}
    \end{subfigure}
    \hfill 
    \begin{subfigure}[b]{0.51\textwidth}
        \includegraphics[trim=20pt 100pt 20pt 80pt, 
            clip,width=\textwidth]{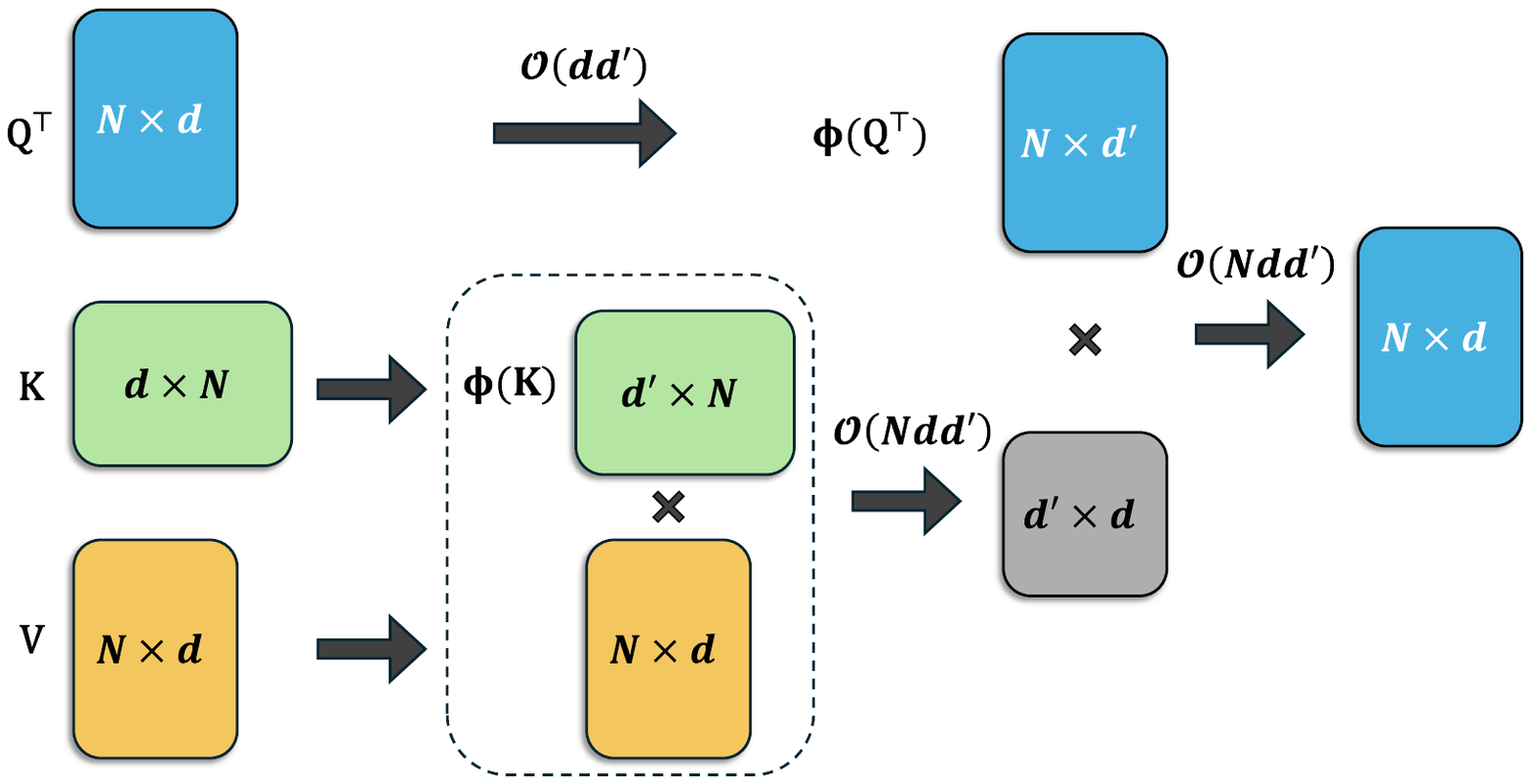}
        \caption{Kernel-based linear attention.}
        \label{cg2}
    \end{subfigure}
    \caption{Comparison between softmax attention and kernel-based linear attention. Softmax attention requires $\mathcal{O}(N^2d)$ time complexity and $\mathcal{O}(N^2+Nd)$ space complexity, whereas kernel-based linear attention achieves $\mathcal{O}(Ndd')$ time complexity and $\mathcal{O}(Nd+Nd'+dd')$ space complexity, where $N,d,d'$ denote input sequence length, hidden dimension, and feature map dimension respectively. When $N \gg d $ and $N \gg d'$, linear attention achieves significant savings in both time and space.}
    \label{attention comparison}
\end{figure*}

Some works are based on sparse attention~\cite{reformer,routing,bigbird}, which modifies the attention patterns by restricting the attention range of each token to focus only on a subset of critical tokens. However, these attention patterns are not rigorously validated and are often based on empirically derived patterns, making it difficult to guarantee their consistently superior performance across diverse data domains. 
The other category is kernel-based linear attention~\cite{lineartrans,performer,cosformer}, which represents the process of computing dot products and applying softmax using kernels.
The kernel can be written as an inner product of feature maps. By rearranging the order of matrix multiplications, the quadratic complexity of attention can be reduced to linear complexity. 
\cref{attention comparison} illustrates the difference of the computational processes between kernel-based linear attention and softmax attention. 

Here, we focus on the kernel-based linear attention whose core lies in constructing an effective feature map. 
While numerous methods aim to recover softmax attention, there is still no proof that softmax is invariably optimal under all circumstances. 
Therefore, a more effective approach is to apply learnable kernels that are directly learned from data. 
However, how to construct sufficiently flexible learnable kernels that guarantee strong performance remains an open problem. 
Hedgehog~\cite{hedgehog} and Polaformer~\cite{polaformer} empirically exploit the low-entropy property of softmax attention by combining learnable mappings with spiky functions such as exponentials and power functions. 
However, these methods do not guarantee that the learned kernel is sufficiently expressive to encompass a wide variety of kernels, which can compromise their performance. 

In this work, we propose a {\bfseries Flex}ible Linear Trans{\bfseries former} with learnable kernel (\textbf{Flexformer}). 
Flexformer extends kernel-based linear attention with random Fourier features by making the feature map fully learnable. 
In contrast to existing learnable kernel methods, Flexformer is able to model a substantially broader and more expressive family of kernels. 
Specifically, spectral representation theory states that a kernel admits an inverse Fourier transform representation in terms of its spectral measure, which can be approximated through Monte Carlo sampling. 
In Flexformer, instead of sampling frequencies from a fixed spectral density, we treat the frequencies as learnable parameters and optimize them directly from data, thereby enabling the attention kernel to be learned in a data-driven manner. 
We propose both stationary and nonstationary versions of Flexformer, with the nonstationary formulation being strictly more expressive. 
Theoretically, we can show that the kernel family of Flexformer contains the softmax kernel. This guarantees that Flexformer can, at a minimum, learn a softmax attention, and potentially learn attention patterns that outperform softmax. 
Flexformer can be trained either from scratch or by distilling~\cite{hedgehog} from a pretrained Transformer. 
We conduct extensive experiments on multiple benchmarks, including language modeling and classification tasks, demonstrating Flexformer's superior performance over existing baselines, as well as its strong capability to recover the softmax kernel and effectively transfer the learned kernel. As a variant of linear attention, it significantly improves efficiency and reduces memory consumption when processing extremely long sequences. 
Specifically, our contributions are summarized as follows: 
\begin{itemize}
    \item We propose Flexformer, which leverages learnable attention kernels constructed via random Fourier features, achieving high expressiveness while preserving linear time and space complexity w.r.t. sequence length. 
    \item We evaluate Flexformer through from-scratch training on language modeling and sequence classification tasks. The results show that Flexformer consistently outperforms other linear-attention baselines, highlighting its strong effectiveness and efficiency. 
    \item Flexformer can faithfully approximate softmax attention through attention weight distillation, enabling efficient linear-attention replacements with preserved performance, and demonstrates strong kernel transferability in cross-domain settings. 
\end{itemize}

\section{Preliminaries}
\label{preliminaries}
In this section, we introduce the background knowledge of self-attention mechanism and kernel-based linear attention. 

\subsection{Self-attention Mechanism}
Given an input sequence $\mathbf{X}=[\text{X}_1,\cdots,\text{X}_N]^\top \in \mathbb{R}^{N\times d}$, where $N$ and $d$ denote the input length and embedding dimension, respectively. Each token $\text{X}_i$ is projected through linear maps to generate the query, key, and value tensors, formally expressed as $\text{Q}_i=\text{W}_Q\text{X}_i, \text{K}_i=\text{W}_K\text{X}_i,\text{V}_i=\text{W}_V\text{X}_i, i=1,\cdots ,N$. For each query tensor, a similarity measure is computed with all key tensors. These similarity scores are then normalized to form attention weights, which are used to perform a weighted summation over the corresponding value tensors, thereby generating context-aware representations. Formally, the output of the self-attention module for the $i$-th token can be expressed as:
\begin{equation}
\label{attention}
    \text{O}_i = \frac{\sum\limits_{j=1}^N \text{sim}(\text{Q}_i, \text{K}_j)\text{V}_j}{\sum\limits_{j=1}^N \text{sim}(\text{Q}_i, \text{K}_j)},
\end{equation}
where $\text{sim}(\cdot,\cdot):\mathbb{R}^{d}\times\mathbb{R}^{d}\to \mathbb{R}_+$ denotes the similarity measure function. In vanilla Transformer~\cite{transformer} with softmax attention, $\text{sim}(\text{Q}_i,\text{K}_j)$ is defined as $\text{sim}(\text{Q}_i,\text{K}_j)=\exp\left(\frac{\text{Q}_i^\top\text{K}_j}{\sqrt{d}}\right)$.

\subsection{Kernel-based Linear Attention}
\label{section linear attention}
If we directly follow the computation order in \cref{attention} by first calculating the similarity between queries and keys at all pairwise positions to obtain a $N\times N$ attention weight matrix, and then multiplying with values, it would result in a time complexity of $\mathcal{O}(N^2d)$ and a space complexity of $\mathcal{O}(N^2+Nd)$, leading to substantial computational and memory overhead as $N$ increases.

To address the aforementioned issues, we assume an attention kernel $k(\mathbf{x},\mathbf{y})$ with a range of $\mathbb{R}_+$ as the similarity function, and let $\phi:\mathbb{R}^d \to \mathbb{R}^{d'}$, where $d'$ is the dimension of the kernel-induced feature space, denote the corresponding feature map, such that: 
\begin{equation}
\label{kernel}
k(\mathbf{x},\mathbf{y})=\phi(\mathbf{x})^\top\phi(\mathbf{y}).
\end{equation}
Then the computation of the self-attention in \cref{attention} can be expressed as:
\begin{equation}
\label{linear att}
    \text{O}_i^\top = \frac{\sum\limits_{j=1}^N \phi(\text{Q}_i)^\top\phi(\text{K}_j)\text{V}_j^\top}{\sum\limits_{j=1}^N \phi(\text{Q}_i)^\top\phi(\text{K}_j)}=\frac{\phi(\text{Q}_i)^\top\sum\limits_{j=1}^N \phi(\text{K}_j)\text{V}_j^\top}{\phi(\text{Q}_i)^\top\sum\limits_{j=1}^N \phi(\text{K}_j)}. 
\end{equation}
This approach avoids explicitly computing the pairwise similarity between any query and key. Instead, we first precompute $\sum_{j=1}^N \phi(\text{K}_i)\text{V}_j^\top$ and $\sum_{j=1}^N \phi(\text{K}_j)$, which are shared across all positional queries, and then multiply them with each positional query $\text{Q}_i$. Thus, the time complexity of this approach is $\mathcal{O}(Ndd')$ and the space complexity is $\mathcal{O}(Nd+Nd'+dd')$, achieving linear scaling w.r.t.  the sequence length. When $N\gg d,d'$, it demonstrates significant efficiency improvements and memory savings compared to standard softmax attention.

\section{Methodology}
In this section, we first review how softmax kernel can be approximated using  random Fourier features. We then extend this formulation to a fully learnable kernel, resulting in the proposed Flexformer. 

\subsection{Softmax Kernel with Random Fourier Features}
In softmax attention, the similarity measure can be expressed as the following kernel. We hereafter refer to this as the softmax kernel: 
\begin{equation*}
    k_{\text{SM}}(\mathbf{x},\mathbf{y})=\exp\left(\frac{\mathbf{x}^\top\mathbf{y}}{\sqrt{d}}\right). 
\end{equation*}
Through Taylor expansion, we can derive feature map $\phi$ satisfying \cref{kernel}. Unfortunately, the dimensionality of $\phi$ is infinite~\cite{attention_inf}. Therefore, we need to seek a finite-dimensional $\phi$ such that the softmax kernel can be effectively approximated. 
The softmax kernel can be decomposed in the following manner: 
\begin{align}
\label{softmax kernel}
    k_{\text{SM}}(\mathbf{x},\mathbf{y})&=\exp\left(\frac{||\mathbf{x}||^2+||\mathbf{y}||^2-||\mathbf{x}-\mathbf{y}||^2}{2\sqrt{d}}\right) =\textcolor{blue}{\exp\left(\frac{||\mathbf{x}||^2}{2\sqrt{d}}\right)}\textcolor{red}{\exp\left(-\frac{||\mathbf{x}-\mathbf{y}||^2}{2\sqrt{d}}\right)}\textcolor{blue}{\exp\left(\frac{||\mathbf{y}||^2}{2\sqrt{d}}\right)}. 
\end{align}
The blue terms depend exclusively on vector norms, enabling all positions to potentially attend strongly to certain keys with large norm magnitudes at critical locations, and they are crucial for preserving the spikiness of the softmax. The red term, which focuses exclusively on a specific similarity between $\mathbf{x}$ and $\mathbf{y}$, is essentially a \emph{Gaussian kernel} with bandwidth $\sigma^2 = \sqrt d$. As a bounded, continuous, and positive definite stationary (a.k.a. translation-invariant) kernel, it can be approximated by the random Fourier features \citep{Randomfeature}. 

\begin{theorem}~\cite{Bochner} 
\label{bochner}
A stationary kernel $k(\mathbf{x},\mathbf{y})=k(\mathbf{x}-\mathbf{y}):\mathbb{R}^{d}\times\mathbb{R}^{d}\to \mathbb{R}$ is a bounded, continuous, and positive definite if and only if it can be represented as:
\[
k(\mathbf{x}-\mathbf{y})=\int_{\mathbb{R}^{d}}\exp(i(\mathbf{\omega}^\top(\mathbf{x}-\mathbf{y}))d\mu(\mathbf{\omega}),
\]
where $\mu(\mathbf{\omega})$ is a bounded non-negative measure associated to the spectral density $p(\omega) = \frac{\mu(\omega)}{\mu(\mathbb{R}^d)}$.
\end{theorem}
Let $\sigma^2 = \mu(\mathbb{R}^d)$ denote the total measure of the space.
Then \cref{bochner} can be expressed in the following expectation form, which can be approximated by Monte Carlo: 
\begin{equation}
\begin{aligned}
\label{phi_n}
    k(\mathbf{x}-\mathbf{y}) 
    =&\sigma^2\mathbb{E}_{p(\omega)}(\cos(\omega^\top(\mathbf{x}-\mathbf{y}))+i\sin(\omega^\top(\mathbf{x}-\mathbf{y}))) \\
    =&\sigma^2\mathbb{E}_{p(\omega)}(\cos\omega^\top\mathbf{x}\cos\omega^\top\mathbf{y}+\sin\omega^\top\mathbf{x}\sin\omega^\top\mathbf{y})  \\
    \approx & \frac {\sigma^2} n \sum_{i=1}^n(\cos\omega_i^\top\mathbf{x}\cos\omega_i^\top\mathbf{y}+\sin\omega_i^\top\mathbf{x}\sin\omega_i^\top\mathbf{y})  \\
    =&\phi_n(\mathbf{x})^\top\phi_n(\mathbf{y}),
\end{aligned}
\end{equation}
where $\phi_n(\mathbf{x})= 
    \frac{\sigma}{\sqrt{n}}[\cos\omega_1^\top\mathbf{x},\cdots,\cos\omega_n^\top\mathbf{x},\sin\omega_1^\top\mathbf{x},\cdots,\sin\omega_n^\top\mathbf{x}]^\top$. 
$\phi_n(\mathbf{x})$ is called Fourier feature, $\{\omega_i\}^n_{i=1}$ are i.i.d. frequencies from $p(\omega)$ and $n$ denotes the number of frequency samples. 
The second equality in the derivation holds because we constrain the range of $k(\mathbf{x}-\mathbf{y})$ to $\mathbb{R}$, which enforces symmetry in the measure $p$ such that $p(\omega) = p(-\omega)$, consequently giving $\mathbb{E}_{p(\omega)}(i\sin(\omega^\top(\mathbf{x}-\mathbf{y})))=0$.

Substituting \cref{phi_n} into the red term of \cref{softmax kernel} yields a random Fourier feature representation of the softmax kernel:
\begin{equation*}
    k_{\text{SM}}(\mathbf{x},\mathbf{y}) \approx \tilde{\phi}_n(\mathbf{x})^\top\tilde{\phi}_n(\mathbf{y}),
\end{equation*}
where $\tilde{\phi}_n(\mathbf{x})=\exp\left(\frac{||\mathbf{x}||^2}{2\sqrt{d}}\right)\phi_n(\mathbf{x})$, $\{\omega_i\}^n_{i=1}$ are i.i.d. frequencies from a \emph{Gaussian spectral density}. 
Based on the above representation, the kernel-based linear attention framework can be employed to reduce the complexity of softmax attention to linear \citep{RFA}. 

The above approach relies on a crucial assumption: it implicitly treats softmax attention as optimal, and therefore fixes the similarity function to the softmax kernel, which is non-learnable. Once the hidden dimension $d$ is specified, the softmax kernel—and consequently its corresponding Gaussian spectral density $p(\omega)$—is fully determined, rendering the resulting attention patterns non-learnable. However, a growing body of work~\cite{hedgehog,polaformer} has shown that softmax attention is not optimal in many scenarios, and that the attention kernel should instead be learned in a data-driven manner. 
This motivates extending the above framework to the learnable attention kernel.


\subsection{Flexformer}
\label{flexformer}
In this section, we first extend the fixed, non-learnable attention kernel to a learnable one, allowing attention patterns to adapt to data and yielding Flexformer. Then we further extend this formulation from stationary kernels to nonstationary kernels, resulting in a nonstationary variant of Flexformer.


\paragraph{Flexformer with learnable stationary kernel.}
As stated in \cref{bochner}, a stationary kernel $k(\mathbf{x}-\mathbf{y})$ is in one-to-one correspondence with its spectral density $p(\omega)$. Consequently, learning a kernel from data is equivalent to learning its spectral density from data.
Therefore, in this work, we directly treat the frequencies $\{\omega_i\}^n_{i=1}$ as learnable parameters and optimize them end-to-end. This is equivalent to learning an optimal spectral density $p(\omega)$, and hence learning an optimal attention kernel. 
Similar techniques have been extensively explored in the Gaussian process literature~\citep{frequency,yaglom2}. 
As \cref{bochner} holds for all stationary kernels satisfying certain conditions, the kernels learned in this manner are sufficiently flexible. 
We refer to this method as $\textbf{Flexformer}_\textbf{s}$, where the subscript $\text{s}$ denotes the stationary kernel. 

It is worth noting that when the learned frequencies $\{\omega_i\}^n_{i=1}$ follow a Gaussian distribution, this yields an approximation of the softmax kernel in \cref{softmax kernel}, analogous to Random Feature Attention~\cite{RFA}.
This observation implies that the kernel family induced by Flexformer strictly contains the unbiased estimate of the softmax kernel. 
As a result, Flexformer is guaranteed to recover softmax attention closely, while also being capable of learning potentially better attention patterns directly from data.

\paragraph{Flexformer with learnable nonstationary kernel.}
The red term in the softmax kernel in \cref{softmax kernel} is stationary, which motivates the use of \cref{bochner} to derive $\text{Flexformer}_{\text{s}}$. However, a stationary kernel implies that its value depends only on the difference 
$\mathbf{x}-\mathbf{y}$, which limits flexibility.
An immediate and natural extension is to replace this stationary kernel with a nonstationary one, thereby enabling the model to learn more diverse attention patterns and adapt more flexibly to the data.
Since \cref{bochner} only characterizes stationary kernels, it cannot be directly applied here. 
Instead, we resort to the Yaglom theorem, which generalizes Bochner’s result to nonstationary kernels.



\begin{theorem}~\cite{Yaglom1} 
\label{yaglom}
A nonstationary kernel $k(\mathbf{x},\mathbf{y}):\mathbb{R}^{d}\times\mathbb{R}^{d}\to \mathbb{R}$ is a bounded,
continuous, and positive definite if and only if it can be represented as:
\[
k(\mathbf{x},\mathbf{y})=\int_{\mathbb{R}^{d}\times\mathbb{R}^{d}}\exp(i(\mathbf{\omega}^\top_1\mathbf{x}-\mathbf{\omega}^\top_2\mathbf{y}))d\mu(\mathbf{\omega}_1,\mathbf{\omega}_2),
\]
where $\mu(\mathbf{\omega}_1,\mathbf{\omega}_2)$ is a Lebesgue-Stieltjes measure associated to some bounded positive semi-definite spectral density $p(\mathbf{\omega}_1,\mathbf{\omega}_2)=\frac{\mu(\mathbf{\omega}_1,\mathbf{\omega}_2)}{\mu(\mathbb{R}^d,\mathbb{R}^d)}$.
\end{theorem}

In fact, the red term in the softmax kernel in \cref{softmax kernel} is indeed a positive definite kernel. 
Therefore, after extending it to the nonstationary setting, we continue to assume that it remains positive definite. 
When approximating the kernel in \cref{yaglom} via Monte Carlo sampling methods, to ensure the positive semi-definiteness of the spectral density $p(\boldsymbol{\omega}_1, \boldsymbol{\omega}_2)$, we first assume that it is symmetrized by an arbitrary joint probability density: 
\begin{align*}
    p(\omega_1,\omega_2)
    =\frac{1}{4}(g(\omega_1,\omega_1)+g(\omega_2,\omega_2)+g(\omega_1,\omega_2)+g(\omega_2,\omega_1)).
\end{align*}
Let $\sigma^2 = \mu(\mathbb{R}^d,\mathbb{R}^d)$ denote the total measure of the space. 
Then \cref{yaglom} can be expressed in the following expectation form, which can be approximated by Monte Carlo: 
\begin{align}
\label{phi_non}
    k(\mathbf{x},\mathbf{y}) \notag
    =&\frac{\sigma^2}{4}\mathbb{E}_{g(\omega_1,\omega_2)}((\cos\omega^\top_1\mathbf{x}+\cos\omega^\top_2\mathbf{x})(\cos\omega^\top_1\mathbf{y}+\cos\omega^\top_2\mathbf{y}) \notag \\
    &+(\sin\omega^\top_1\mathbf{x}+\sin\omega^\top_2\mathbf{x})(\sin\omega^\top_1\mathbf{y}+\sin\omega^\top_2\mathbf{y})) \notag \\
    \approx &\frac{\sigma^2}{4n}\sum_{i=1}^n((\cos\omega^\top_{1i}\mathbf{x}+\cos\omega^\top_{2i}\mathbf{x})(\cos\omega^\top_{1i}\mathbf{y}+\cos\omega^\top_{2i}\mathbf{y}) \notag \\
    &+(\sin\omega^\top_{1i}\mathbf{x}+\sin\omega^\top_{2i}\mathbf{x})(\sin\omega^\top_{1i}\mathbf{y}+\sin\omega^\top_{2i}\mathbf{y})) \notag \\
    =&\frac{\sigma^2}{4n}\sum_{i=1}^n(\cos\omega'^\top_{1i}\mathbf{x}\cos\omega'^\top_{2i}\mathbf{x}\cos\omega'^\top_{1i}\mathbf{y}\cos\omega'^\top_{2i}\mathbf{y} +\sin\omega'^\top_{1i}\mathbf{x}\cos\omega'^\top_{2i}\mathbf{x}\sin\omega'^\top_{1i}\mathbf{y}\cos\omega'^\top_{2i}\mathbf{y})
    \notag \\
    =&\phi_n(\mathbf{x})^\top\phi_n(\mathbf{y}),
\end{align}
where
   $\phi_n(\mathbf{x})=\frac{\sigma}{2\sqrt{n}}[\cos\omega'^\top_{11}\mathbf{x}\cos\omega'^\top_{21}\mathbf{x},\cdots,\cos\omega'^\top_{1n}\mathbf{x}\cos\omega'^\top_{2n}\mathbf{x},
\sin\omega'^\top_{11}\mathbf{x}\cos\omega'^\top_{21}\mathbf{x},\cdots,\sin\omega'^\top_{1n}\mathbf{x} \\ \cos\omega'^\top_{2n}\mathbf{x}]^\top$,
$\{(\omega_{1i}, \omega_{2i})\}_{i=1}^n$ are i.i.d. frequencies from $g(\omega_1,\omega_2)$, $n$ denotes the number of frequencies, and
    $\omega_{1i}'= \frac{\omega_{1i}+\omega_{2i}}{2}, \omega_{2i}'= \frac{\omega_{1i}-\omega_{2i}}{2}$.
Substituting \cref{phi_non} into the red term of \cref{softmax kernel} yields a random Fourier feature representation of the attention kernel. 
For the blue term $\exp\!\left(\frac{\lVert \mathbf{x} \rVert^2}{2\sqrt{d}}\right)$, we preserve its functional form while replacing the fixed scaling factor $2\sqrt{d}$ with a learnable parameter $\exp(\tau)$. 
This leads to the following attention kernel approximation:
\begin{equation*}
    k(\mathbf{x}, \mathbf{y}) \approx \tilde{\phi}_n(\mathbf{x})^\top \tilde{\phi}_n(\mathbf{y}),
\end{equation*}
where $\tilde{\phi}_n(\mathbf{x}) = \exp\!\left(\frac{\lVert \mathbf{x} \rVert^2}{\exp(\tau)}\right)\phi_n(\mathbf{x})$. 
Both scaling parameter $\tau$ and spectral pairs $\{(\omega_{1i}, \omega_{2i})\}_{i=1}^n$ are treated as learnable parameters. 
We refer to this nonstationary variant as $\textbf{Flexformer}_\textbf{n}$, where the subscript $\text{n}$ denotes the nonstationary kernel. 
We provide an analysis of the time and space complexity, as well as the additional parameter overhead of Flexformer, in Appendix~\ref{complexity}.

\section{Experiments}

To evaluate the effectiveness and efficiency of Flexformer as a linear-attention variant, we conduct extensive experiments on long-context classification, autoregressive language modeling, softmax attention recovery, and cross-domain kernel transfer. All results report the mean over three runs, with full results including standard deviations provided in Appendix~\ref{std}. Additional hyperparameter sensitivity analyses, and experimental results are deferred to Appendix~\ref{app_d}. 

\subsection{Long Sequence Classification}

\begin{table}[t]
\centering
\caption{Performance comparison on LRA benchmarks. 
All reported metrics represent test accuracy (\%), where higher values are better. 
The best results are in \colorbox{lightblue}{\textbf{bold}}, 
and the second best in \colorbox{lightgreen}{\underline{underlined}}.}
\label{lra result}
\resizebox{0.8\textwidth}{!}{
\begin{tabular}{lcccccc}
\toprule
Model 
& ListOps & Text & Retrieval & Image & Pathfinder & Average \\
\midrule
Transformer~\cite{transformer} 
& \colorbox{lightgreen}{\underline{37.23}} & 65.66 & 64.49 & 41.89 & 73.46 & 56.55 \\
Reformer~\cite{reformer} 
& 36.62 & 63.52 & 57.60 & 40.06 & 73.61 & 54.28 \\
Longformer~\cite{longformer} 
& 36.46 & 63.09 & 59.24 & 41.23 & 72.44 & 54.49 \\
Linear Trans.~\cite{lineartrans} 
& 17.42 & 62.64 & 58.92 & 43.40 & 75.11 & 51.49 \\
BigBird~\cite{bigbird} 
& 36.88 & 65.34 & 64.23 & 41.93 & 73.58 & 56.39 \\
Performer~\cite{performer} 
& 16.97 & 63.67 & 55.17 & 41.23 & 75.38 & 50.48 \\
RFA~\cite{RFA} 
& 17.20 & 64.89 & 58.76 & 41.00 & 74.51 & 51.27 \\
Skyformer~\cite{Skyformer} 
& 36.71 & 65.51 & 68.05 & 41.36 & 74.18 & 57.16 \\
Cosformer~\cite{cosformer} 
& \colorbox{lightblue}{\textbf{37.44}} & 65.16 & 64.04 & \colorbox{lightgreen}{\underline{43.56}} & 72.32 & 56.50 \\
Hedgehog~\cite{hedgehog} 
& 37.02 & 64.10 & 68.67 & 41.64 & 75.89 & 57.47 \\
Polaformer~\cite{polaformer} 
& 36.93 & \colorbox{lightblue}{\textbf{69.65}} & 65.39 & 40.95 & 73.26 & 57.23 \\
\midrule
$\text{Flexformer}_\text{s}$ (Ours) 
& 37.09 & \colorbox{lightgreen}{\underline{67.38}} & \colorbox{lightgreen}{\underline{71.55}} & 43.39 & \colorbox{lightgreen}{\underline{76.15}} & \colorbox{lightgreen}{\underline{59.11}} \\
$\text{Flexformer}_\text{n}$ (Ours) 
& 37.18 & 65.72 & \colorbox{lightblue}{\textbf{73.94}} & \colorbox{lightblue}{\textbf{45.42}} & \colorbox{lightblue}{\textbf{77.66}} & \colorbox{lightblue}{\textbf{59.99}} \\
\bottomrule
\end{tabular}}
\end{table}

To evaluate Flexformer’s ability to model long sequences and reduce cost, we conduct long-sequence classification experiments on the Long Range Arena (LRA) benchmark~\cite{LRA}. LRA is designed to assess both the effectiveness and efficiency of sequence models under long-context settings. It includes tasks such as Long ListOps~\cite{listops}, byte-level text classification~\cite{imdb}, byte-level document retrieval~\cite{aan}, image classification on sequences of pixels~\cite{cifar10}, and Pathfinder~\cite{pathfinder}. The input sequence lengths in these tasks range from $1\text{K}$ to $4\text{K}$. We adopt the same hyperparameter configurations as those used in the official open-source implementation. 

We present in \cref{lra result} a comprehensive performance comparison between Flexformer, the vanilla Transformer, several classical Transformer variants, and a range of existing linear attention methods. 
Early approaches that rely on random Fourier features to obtain unbiased estimators of the softmax kernel, such as RFA and Performer, substantially underperform the vanilla Transformer, particularly on challenging tasks such as ListOps and document retrieval. 
Methods based on learnable kernels, including Hedgehog and Polaformer, exhibit strong performance on certain specific tasks, but their effectiveness does not consistently generalize across all benchmarks.
In contrast, the proposed Flexformer, benefiting from its flexible and expressive feature map representation, achieves consistently strong performance across all tasks. 
In particular, $\text{Flexformer}_{\text{n}}$ attains the best results on document retrieval, image classification, and Pathfinder, while remaining highly competitive on the remaining tasks. 
Overall, it achieves the highest average accuracy, corresponding to a $4.4\%$ relative improvement over the best existing linear attention method. 
The stationary variant, $\text{Flexformer}_{\text{s}}$, ranks second overall and also outperforms all prior linear attention approaches.

\begin{figure*}[t]
    \centering
    \begin{subfigure}[b]{0.66\linewidth}
        \includegraphics[trim=2pt 5pt 2pt 2pt, 
            clip,width=\textwidth]{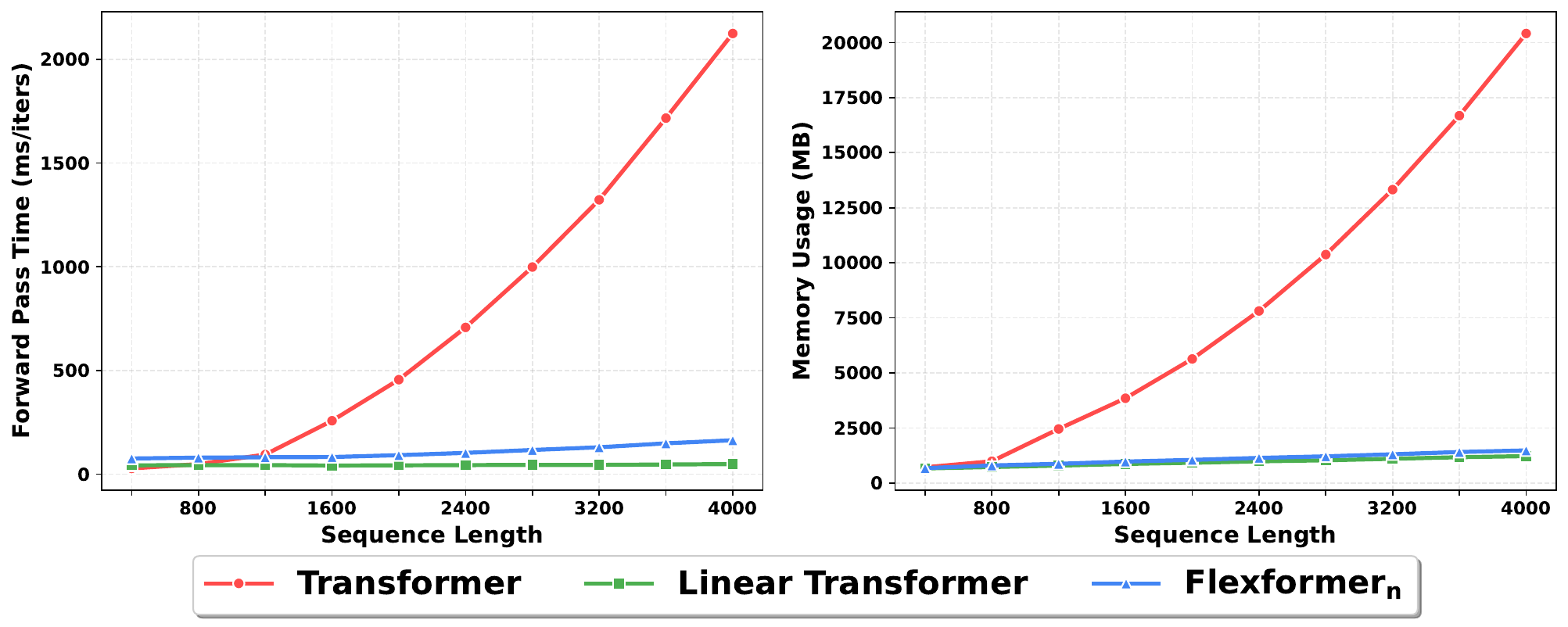}
        \caption{Forward Pass Time \& GPU Memory Usage vs. Sequence Length.}
        \label{linear complexity}
    \end{subfigure}
    \hfill 
    \begin{subfigure}[b]{0.33\linewidth}
        \includegraphics[
            trim=50pt 50pt 50pt 50pt,  
            clip,
            width=\textwidth
        ]{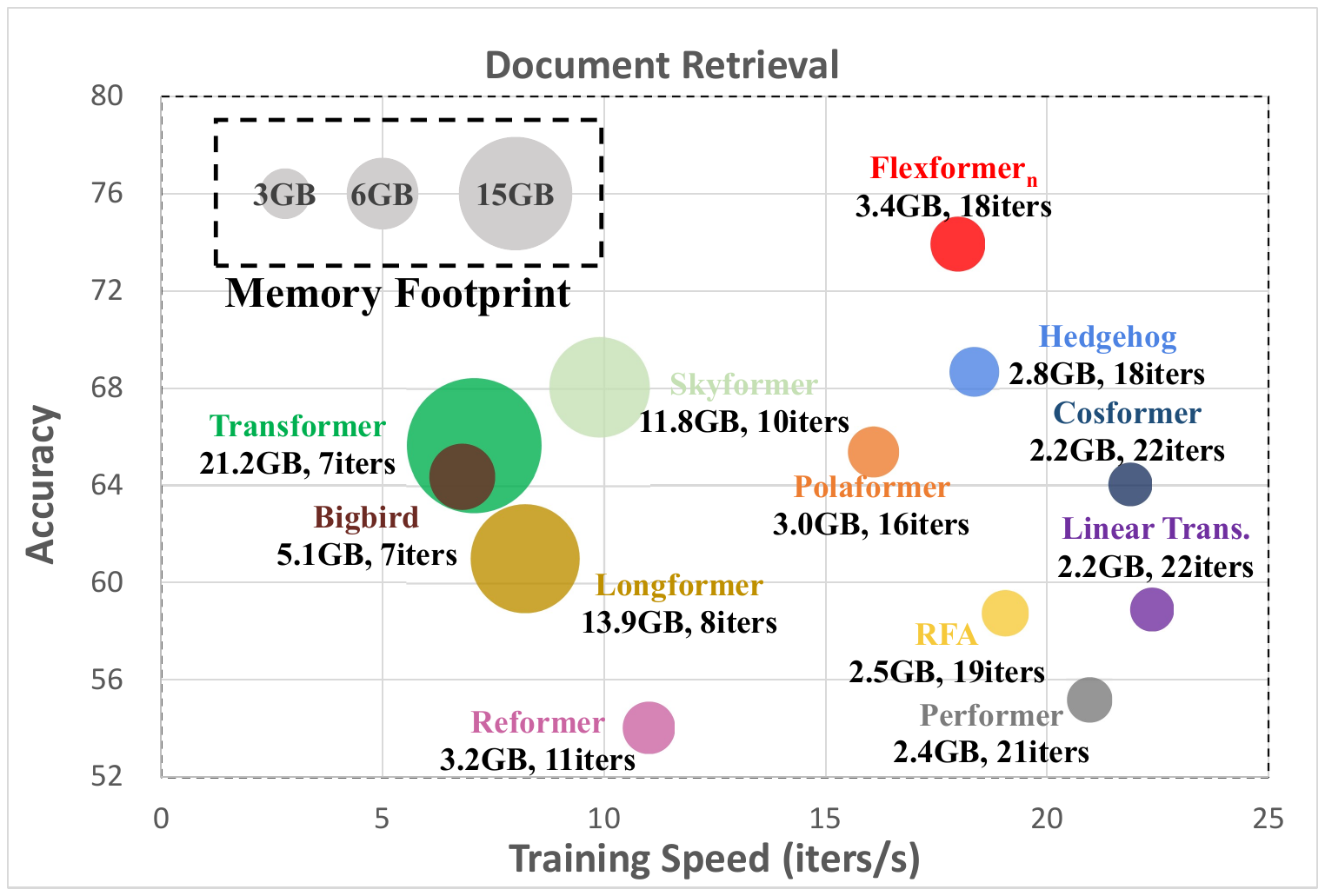}
        \caption{Test accuracy, training speed and peak GPU memory usage comparison on document retrieval.}
        \label{eff aan}
    \end{subfigure}
    \caption{Comparison of runtime and memory usage across models. As shown in (a), Flexformer scales linearly with sequence length in both time and memory. Together with (b), the results show that Flexformer achieves higher efficiency and lower memory usage than other baselines.}
    \label{efficiency comparison}
\end{figure*}

We further compare the computational time and memory usage of different Transformer variants on the LRA benchmark in \cref{efficiency comparison}. 
In \cref{linear complexity}, we evaluate the forward-pass computation time and memory consumption of the vanilla Transformer, Linear Transformer, and our proposed $\text{Flexformer}_{\text{n}}$ under same hyperparameter configurations on the document retrieval task using input sequences of varying lengths. 
As expected, both the computation time and memory usage of the vanilla Transformer scale quadratically with the sequence length, resulting in a rapid increase as the input length grows. 
In contrast, $\text{Flexformer}_{\text{n}}$ exhibits linear complexity with respect to sequence length, closely matching the scaling behavior of the Linear Transformer.

In \cref{eff aan}, we compare different Transformer variants under identical training hyperparameter settings in terms of test accuracy, training throughput, and peak memory usage on the document retrieval task. 
This configuration uses a maximum sequence length of $N=4000$ and a per-head dimension of $d=64$ in the encoder, satisfying $N \gg d$. 
Under this setting, all linear attention models demonstrate substantially faster training speed and significantly lower memory consumption compared to the vanilla Transformer. 
Compared with other linear attention methods, $\text{Flexformer}_{\text{n}}$ introduces additional parameters and employs a slightly more complex feature map, leading to marginally reduced efficiency and memory savings. 
However, this modest overhead is offset by significantly improved model performance while preserving linear time and space complexity. 
On the document retrieval task, Flexformer achieves a $2.6\times$ training speedup and an $84\%$ reduction in memory usage relative to the vanilla Transformer. 


\subsection{Autoregressive Language Modeling}

To evaluate Flexformer’s effectiveness on text generation tasks, we conduct autoregressive language modeling experiments on WikiText-103~\cite{wikitext}. 
We use the vanilla Transformer with adaptive inputs \cite{adaptive} as the base architecture and replace its softmax attention with various linear attentions. 
Following the protocol of \cite{RFA}, we evaluate small and large models with roughly 41M and 247M parameters, respectively. 
The detailed hyperparameter settings are provided in Appendix~\ref{detail lm}. 
For all experiments, we set the sentence block size to 512 and train the models for 100K steps using a batch size of 128.  

The final evaluation results of all baselines on both the development and test sets are reported in \cref{wikitext result}. 
Under both model configurations, the vanilla Transformer with softmax attention consistently achieves strong performance. 
Hedgehog and Polaformer enable linear attention to preserve the low-entropy property of softmax attention, resulting in substantial gains over earlier linear-attention methods. 
Nevertheless, they still fall short of matching the vanilla Transformer, although the performance gap narrows under the large model configuration.

\begin{wraptable}{r}{0.49\textwidth}
\vspace{-0.45cm}
\caption{Language model perplexity (PPL) on the WikiText-103 development and test sets. 
Lower PPL is better. 
The best results for each benchmark are in \colorbox{lightblue}{\textbf{bold}}, 
and the second best in \colorbox{lightgreen}{\underline{underlined}}.}
\label{wikitext result}
\resizebox{\linewidth}{!}{
\begin{tabular}{lcccc}
\toprule
 & \multicolumn{2}{c}{\makecell{Small \\ ($41\text{M}$ Params)}} 
 & \multicolumn{2}{c}{\makecell{Big \\ ($247\text{M}$ Params)}} \\
\cmidrule(lr){2-3} \cmidrule(lr){4-5}
Model & Dev. & Test & Dev. & Test \\
\midrule 
Vanilla Trans.          
& \colorbox{lightblue}{\textbf{30.0}} 
& \colorbox{lightblue}{\textbf{31.3}} 
& 22.6 & 23.8 \\
\midrule
Linear Trans. & 34.1 & 35.6 & 27.9 & 29.2 \\
RFA           & 33.8 & 35.3 & 26.7 & 28.0 \\
Cosformer     & 32.8 & 34.2 & 24.9 & 26.2 \\
Hedgehog      & 31.7 & 33.2 & 23.0 & 24.2 \\
Polaformer    & 31.5 & 32.9 & 23.1 & 24.3 \\
\midrule          
$\text{Flexformer}_\text{s}$ 
& 31.1 & 32.7 
& \colorbox{lightgreen}{\underline{22.2}} 
& \colorbox{lightgreen}{\underline{23.4}} \\
$\text{Flexformer}_\text{n}$ 
& \colorbox{lightgreen}{\underline{30.3}} 
& \colorbox{lightgreen}{\underline{31.6}} 
& \colorbox{lightblue}{\textbf{21.2}} 
& \colorbox{lightblue}{\textbf{22.3}} \\
\bottomrule
\end{tabular}}
\vspace{-1cm}
\end{wraptable}

Notably, both Flexformer variants outperform all other linear attention methods under the small model configuration and achieve performance very close to that of the vanilla Transformer. 
Under the big model configuration, Flexformer further surpasses the vanilla Transformer, with $\text{Flexformer}_\text{n}$ demonstrating a particularly pronounced improvement. 
This observation suggests that learnable kernels benefit more from increased model capacity, as the additional parameters introduced by kernel learning allow better utilization of larger model scales. 
Importantly, the relative increase in parameter count remains nearly constant across configurations (see Appendix~\ref{detail lm}). 
Moreover, $\text{Flexformer}_\text{s}$ introduces a comparable number of additional parameters to Hedgehog and fewer than Polaformer, yet still achieves superior performance.
 

\subsection{Recovering Softmax Attention}
\label{recover}

Besides training from scratch, another scenario that demands efficient linear attention is when we already have a traditional Transformer pretrained or finetuned, and we need to replace its softmax attention with linear attention while recovering the original performance. We adopt the attention weight distillation loss~\cite{hedgehog} to supervise the attention weights computed by Flexformer with the attention weights derived from softmax. We first download a pretrained 125M RoBERTa~\cite{roberta} model using the fairseq~\cite{fairseq}
library and then finetune it on each dataset of GLUE~\cite{GLUE} benchmark. 
We compare the following categories of methods: (1) Directly finetuning the pretrained RoBERTa on the corresponding dataset. (2) Directly replace the attention computation in RoBERTa with Performer and RFA (they are not learnable), and then finetune the models. (3) Starting from a finetuned RoBERTa model, we perform attention weights distillation on the corresponding dataset to convert it into $\text{Flexformer}_\text{n}$ and Hedgehog, respectively, followed by finetuning of the resulting models. More detailed experimental settings are provided in Appendix~\ref{detail dist}.
For these three methods, we use the same finetuning hyperparameter configuration and present the experimental results in \cref{glue result}. 

\begin{table}[t]
\centering
\caption{Performance comparison on GLUE. On CoLA, the Matthews correlation coefficient is reported; 
on STS-B, the average of Pearson and Spearman correlation coefficients is reported; 
for all other tasks, accuracy is reported. Higher values indicate better performance for all metrics. 
The best results of each task (excluding RoBERTa) are in \colorbox{lightblue}{\textbf{bold}}, 
and the second best in \colorbox{lightgreen}{\underline{underlined}}.}
\label{glue result}
\resizebox{\textwidth}{!}{
\begin{tabular}{lccccccccc}
\toprule
Model & CoLA & SST-2 & MRPC & STS-B & QQP & MNLI & QNLI & RTE & Average \\
\midrule
\rowcolor{yellow!30}
RoBERTa (reference) 
& 61.9 & 94.8 & 88.5 & 90.2 & 91.7 & 87.6 & 92.7 & 76.9 & 85.5 \\
\midrule
Performer (unlearnable) 
& 22.8 & 88.9 & 66.3 & 38.8 & 84.4 & 77.5 & 76.3 & 51.6 & 63.3 \\
RFA (unlearnable) 
& 27.6 & 90.1 & 64.2 & 41.2 & 86.3 & 80.8 & 75.8 & 53.0 & 64.9 \\
Hedgehog (distilled) 
& \colorbox{lightgreen}{\underline{62.5}} 
& \colorbox{lightgreen}{\underline{93.3}} 
& \colorbox{lightblue}{\textbf{86.0}} 
& \colorbox{lightgreen}{\underline{87.7}} 
& \colorbox{lightblue}{\textbf{91.2}} 
& \colorbox{lightgreen}{\underline{83.7}} 
& \colorbox{lightblue}{\textbf{91.2}} 
& \colorbox{lightgreen}{\underline{71.0}} 
& \colorbox{lightgreen}{\underline{83.3}} \\
$\text{Flexformer}_\text{n}$ (distilled) 
& \colorbox{lightblue}{\textbf{63.3}} 
& \colorbox{lightblue}{\textbf{93.6}} 
& \colorbox{lightgreen}{\underline{85.8}} 
& \colorbox{lightblue}{\textbf{88.3}} 
& \colorbox{lightgreen}{\underline{91.1}} 
& \colorbox{lightblue}{\textbf{85.1}} 
& \colorbox{lightgreen}{\underline{91.0}} 
& \colorbox{lightblue}{\textbf{71.5}} 
& \colorbox{lightblue}{\textbf{83.7}} \\
\bottomrule
\end{tabular}}
\vspace{-5pt}
\end{table}

Compared to directly finetuning pretrained RoBERTa, replacing its softmax attention with linear attention methods that provide unbiased estimates of the softmax kernel severely degrades performance, especially on several datasets. In contrast, distilled Hedgehog and Flexformer both nearly fully recover the softmax attention behavior. Furthermore, the learnable kernel demonstrates performance that even surpasses standard softmax attention on CoLA.



\subsection{Transferability of Learned Kernel}
\label{transfer}

In this section, we examine whether the attention kernels learned by Flexformer on one dataset can be transferred to datasets from other domains, thus maintaining strong generalization performance.
To this end, we first perform attention weight distillation on a pretrained RoBERTa model using a single source dataset, converting it into a linear-attention model. Specifically, we use four GLUE datasets—STS-B, MNLI, QQP, and QNLI—to obtain four distilled models, each learned from a different source domain.
We then fine-tune these four distilled models on the remaining GLUE datasets, enabling us to evaluate their cross-domain transferability.
The complete results are in \cref{full glue result}. 
In \cref{transfer result}, we report the averaged performance, where each column shows the average performance on the target task over kernels learned from different source datasets. 

\begin{table}[t]
\centering
\caption{Comparison of the transfer performance of kernels learned by $\text{Flexformer}_\text{n}$ and Hedgehog. 
For each target task, we report the average performance over kernels 
learned from different source domains. 
The best results of each task are in \colorbox{lightblue}{\textbf{bold}}.}
\label{transfer result}
\resizebox{0.87\textwidth}{!}{
\begin{tabular}{lcccccccc}
\toprule
Model & CoLA & SST-2 & MRPC & STS-B & QQP & MNLI & QNLI & RTE \\
\midrule
Hedgehog 
& 55.6 & 92.6 & 76.3 & 85.3 & 89.6 & 82.7 & 87.5 & 53.1 \\
$\text{Flexformer}_\text{n}$ 
& \colorbox{lightblue}{\textbf{60.3}} 
& \colorbox{lightblue}{\textbf{93.3}} 
& \colorbox{lightblue}{\textbf{79.5}} 
& \colorbox{lightblue}{\textbf{86.0}} 
& \colorbox{lightblue}{\textbf{90.7}} 
& \colorbox{lightblue}{\textbf{84.8}} 
& \colorbox{lightblue}{\textbf{89.5}} 
& \colorbox{lightblue}{\textbf{56.1}} \\
\bottomrule
\end{tabular}}
\vspace{-5pt}
\end{table}

We observe that, when transferring the learned attention kernel from a source domain to a target domain, $\text{Flexformer}_{\text{n}}$ consistently achieves superior performance across all settings.
Moreover, compared to the in-domain results reported in \cref{glue result}, Hedgehog suffers a particularly severe performance degradation on the CoLA dataset under kernel transfer, whereas $\text{Flexformer}_{\text{n}}$ exhibits a substantially smaller performance drop.
These results demonstrate that the attention kernels learned by Flexformer possess stronger cross-domain transferability than those of existing baselines.

\section{Related Works}
\label{related}

Kernel-based linear attention methods reduce the quadratic complexity of softmax attention by expressing the attention kernel as an inner product of feature maps. Existing approaches differ in how these feature maps are constructed. Early methods approximate the exponential function via Taylor expansion~\cite{basetaylor}, but even low-order approximations lead to prohibitively high feature dimensions~\cite{taylersoftmax}. Other works adopt simple element-wise mappings, such as Linear Transformer~\cite{lineartrans} and Cosformer~\cite{cosformer}, which are efficient but limited in expressiveness. Random-feature-based methods, including Performer~\cite{performer} and RFA~\cite{RFA}, use random Fourier features to obtain unbiased estimators of the softmax kernel, assuming softmax attention is optimal and fixing the kernel a priori. Recent approaches introduce learnable feature maps: Hedgehog~\cite{hedgehog} exploits the low-entropy property of softmax using exponential-based mappings, while PolaFormer~\cite{polaformer} incorporates polarity-aware mixing and power functions. These methods empirically design kernels with desirable properties, but the expressiveness of their kernel families is not explicitly characterized. In contrast, Flexformer is grounded in spectral representation theory, which establishes a one-to-one correspondence between a kernel and its spectral measure. By treating spectral frequencies as learnable parameters and optimizing them directly from data, Flexformer enables a principled, data-driven construction of attention kernels and can model a broader and more expressive kernel family.

\section{Conclusions}
We proposed Flexformer, a flexible kernel-based linear attention framework that learns attention kernels directly from data using Fourier features. By treating spectral frequencies as trainable parameters, Flexformer expands the expressiveness of linear attention while preserving linear time and space complexity with respect to sequence length. We further introduced stationary and nonstationary variants, depending on the desired level of flexibility in the kernel.
Empirically, extensive experiments on language modeling and sequence classification tasks demonstrate that Flexformer consistently outperforms existing linear attention baselines. Moreover, when distilled from pretrained Transformers, Flexformer can approximate softmax attention while enabling efficient linear attention inference. We also observe strong cross-domain kernel transferability. Overall, Flexformer provides an effective and scalable alternative to softmax attention for long-sequence modeling.

\textbf{Limitations.} Flexformer assumes the learned attention kernel is positive definite, but whether positive definite kernels are truly the best choice for attention remains theoretically unexplored. Establishing a rigorous justification for this assumption is an important direction for future work.

\section*{Acknowledgements}
This work was supported by the NSFC Project (No.62576346), the MOE Project of Key Research
Institute of Humanities and Social Sciences (22JJD110001), the fundamental research funds for the
central universities, and the research funds of Renmin University of China (24XNKJ13), and Beijing
Advanced Innovation Center for Future Blockchain and Privacy Computing.

\bibliographystyle{plain}
\bibliography{reference}

\newpage
\appendix

\section{Experimental Details}

\subsection{Introduction of Datasets}
Statistics on the sizes of all datasets used in our experiments are provided in \cref{dataset stat}. Next, we provide a brief overview of these datasets.

\begin{table}[h]
\centering
\caption{Statistics of the datasets used in our experiments.}
\label{dataset stat}
\small
\begin{tabular}{lccc}
\toprule
Dataset & Train & Valid & Test \\
\midrule
ListOps    & 96K  & 2K   & 2K \\
IMDB       & 25K  & --   & 25K \\
AAN        & 147K & 18K  & 17K \\
CIFAR-10   & 45K  & 5K   & 10K \\
Pathfinder & 160K & 20K  & 20K \\
\midrule
WikiText-103 & 103M & 218K & 246K \\
\midrule
CoLA & 8.5K & -- & 1K \\
SST-2 & 67K & -- & 1.8K \\
MRPC & 3.7K & -- & 1.7K \\
STS-B & 7K & -- & 1.4K \\
QQP & 364K & -- & 391K \\
MNLI & 393K & -- & 20K \\
QNLI & 105K & -- & 5.4K \\
RTE & 2.5K & -- & 3K \\
WNLI & 634K & -- & 146 \\
\bottomrule
\end{tabular}
\end{table}

\subsubsection{Long Range Arena}
Next, we provide a detailed introduction to the datasets of the five tasks in the LRA.
\paragraph{Long ListOps}  A 10-class classification task based on a longer variant of the synthetic ListOps dataset~\cite{listops}. The sequences with a max length of 2K consist of nested expressions using operators like MAX, MIN, and MEDIAN, and models must predict the final result (an integer from 0 to 9), testing their ability to model hierarchical structure over long contexts.

\paragraph{Byte-level Text Classification} A binary classification task using IMDb movie reviews~\cite{imdb} processed as raw byte sequences with fixed length of 4K. The goal is to classify each review as positive or negative sentiment without word segmentation or pretraining.

\paragraph{Byte-level Document Retrieval} A binary classification task based on the ACL Anthology Network (AAN)~\cite{aan} dataset, where models determine whether two 4K-byte academic documents share a citation link, using a two-tower encoding architecture.

\paragraph{Image Classification on Sequences of Pixels} A 10-class classification task using grayscale CIFAR-10 images~\cite{cifar10}, flattened into 1,024-pixel sequences. Models classify each image into one of ten object categories using only sequential pixel values.

\paragraph{Pathfinder} A binary classification task derived from the Pathfinder dataset~\cite{pathfinder} introduced in cognitive psychology and adapted for machine learning. It uses 32×32 synthetic images (1,024 pixels) containing two marked points and dashed paths; the model must determine whether the points are connected by a continuous path, evaluating its capacity for long-range spatial reasoning.

\subsubsection{Wikitext-103}
WikiText-103~\cite{wikitext} is a large-scale language modeling benchmark introduced in this paper to address the limitations of smaller datasets like Penn Treebank. It is constructed from 28,475 high-quality English Wikipedia articles (including 23,805 "Good" and 4,790 "Featured" articles), resulting in a training set of over 103 million words. The dataset preserves original casing, punctuation, and named entities, making it more realistic for real-world applications. After preprocessing using the Moses tokenizer and replacing words occurring fewer than 3 times with <unk>, it has a vocabulary size of 267,735 and a very low out-of-vocabulary rate of $0.4\%$. Unlike shuffled corpora, WikiText-103 maintains full article structure, enabling the evaluation of models on long-range dependencies and rare word prediction.

\subsubsection{GLUE}
We introduce below the eight datasets in GLUE corresponding to their respective tasks.

\paragraph{CoLA (Corpus of Linguistic Acceptability)} The task is to judge whether a sentence is grammatically acceptable in English. The data is drawn from books and journal articles on linguistic theory. Labels are binary: acceptable or unacceptable.

\paragraph{SST-2 (Stanford Sentiment Treebank)} The task is binary sentiment classification of sentences from movie reviews. The data comes from movie reviews. Labels are positive or negative sentiment.

\paragraph{MRPC (Microsoft Research Paraphrase Corpus)} The task is to determine if two sentences are semantically equivalent (paraphrases). Sentence pairs are automatically extracted from online news sources. Labels are binary: paraphrase or not paraphrase.

\paragraph{STS-B (Semantic Textual Similarity Benchmark)} The task is to predict the degree of semantic similarity between two sentences. The data is drawn from news headlines, video/image captions, and natural language inference datasets. Labels are continuous similarity scores ranging from 1 to 5 (a regression task).

\paragraph{QQP (Quora Question Pairs)} The task is to determine whether two questions from the Quora community QA platform are semantically equivalent (duplicates). Labels are binary: duplicate or not duplicate.

\paragraph{MNLI (Multi-Genre Natural Language Inference)} The task is to classify the relationship between a premise and a hypothesis sentence as entailment, contradiction, or neutral. Premise sentences are gathered from ten different sources, including transcribed speech, fiction, and government reports. Labels are three-way: entailment, contradiction, or neutral.

\paragraph{QNLI (Question Natural Language Inference)} This task is derived from the SQuAD question-answering dataset. It is recast as an NLI problem to determine if a sentence from a Wikipedia paragraph contains the answer to a given question. Labels are binary: entailment (contains answer) or not entailment.

\paragraph{RTE (Recognizing Textual Entailment)} The task is to recognize textual entailment in a binary setting. The data combines examples from the RTE1, RTE2, RTE3, and RTE5 challenges, based on news and Wikipedia text. Labels are binary: entailment or not entailment.

\subsection{Setup}
\label{setup}
All of our experiments are implemented using PyTorch 2.9.1~\cite{Pytorch} and conducted on a single Nvidia RTX5090 GPU with 32GB memory. We use the Adam~\cite{adam} optimizer for all experiments.

\subsubsection{Autoregressive Language Modeling}
\label{detail lm}
The detailed hyperparameter configurations for the autoregressive language modeling experiments are listed in \cref{wiki params}. Total number of parameters refers to the Transformer architecture without any additional parameters introduced. Methods with learnable feature maps have a total number of parameters slightly larger than this value. Their specific amounts of additional parameters are shown in \cref{add params}. As discussed in \cref{complexity}, the additional parameters introduced by Flexformer are minimal, all less than $1\%$. The additional parameters of $\text{Flexformer}_\text{n}$ are almost exactly twice those of Hedgehog, whereas $\text{Flexformer}_\text{s}$ introduces nearly the same amount of extra parameters as Hedgehog. Polaformer, in contrast, incurs a larger number of additional parameters because it applies a linear projection after concatenating all heads, rather than using separate projections for each head. In large language models, a greater number of parameters often leads to better performance. Our results show that Flexformer achieves a higher performance gain per additional parameter compared to other models.

\begin{table}[htbp]
\centering
\caption{Detailed hyperparameter settings for autoregressive language modeling experiments.}
\label{wiki params}
\small
\begin{tabular}{lcc}
\toprule
Hyperparameter & Small & Big \\
\midrule
Embedding Size & 512 & 1024 \\
Layers & 6 & 16 \\
Heads & 8 & 16 \\
Head Size & 64 & 64 \\
FFN Size & 2048 & 4096 \\
Batch Size & 128 & 128 \\
Learning Rate & $5\times10^{-4}$ & $5\times10^{-4}$ \\
Total Steps & 100{,}000 & 100{,}000 \\
Warmup Steps & 4{,}000 & 4{,}000 \\
Dropout & 0.1 & 0.2 \\
Gradient Clipping Norm & 0.1 & 0.1 \\
Total Number of Parameters & $41\,\text{M}$ & $247\,\text{M}$ \\
\bottomrule
\end{tabular}
\end{table}

\begin{table}[htbp]
\centering
\caption{Comparison of parameter counts between methods with learnable feature maps and the Vanilla Transformer. ``Relative Parameter Increase'' denotes the percentage increase in model parameters compared to a Vanilla Transformer under the same configuration.}
\label{add params}
\small
\begin{tabular}{lcccc}
\toprule
 & \multicolumn{2}{c}{Small} & \multicolumn{2}{c}{Big} \\
\cmidrule(lr){2-3} \cmidrule(lr){4-5}
Model & Total Params & Rel. Increase (\%) & Total Params & Rel. Increase (\%) \\
\midrule
Transformer & 41M & 0.00 & 247M & 0.00 \\
Hedgehog & 41M & 0.49 & 248M & 0.43 \\
Polaformer & 43M & 3.83 & 262M & 6.74 \\
$\text{Flexformer}_\text{s}$ & 41M & 0.49 & 248M & 0.43 \\
$\text{Flexformer}_\text{n}$ & 42M & 0.97 & 249M & 0.86 \\
\bottomrule
\end{tabular}
\end{table}

\subsection{Recovering Softmax Attention and Transferability of Learned Kernel}
\label{detail dist}
For each dataset of GLUE, we use the preprocessing script  from the Fairseq repository\footnote{\url{https://github.com/facebookresearch/fairseq/blob/main/examples/roberta/preprocess_GLUE_tasks.sh}} to further split the training set into training and validation subsets. The downloaded pre-trained 125M RoBERTa model uses a GPT-style BPE tokenizer with a vocabulary size of approximately 50K, an embedding dimension of 512, 12 encoder layers, and 8 attention heads. All methods use identical settings during finetuning, and both Hedgehog and Flexformer employ the same configuration for distillation.
\paragraph{Finetuning Setting.} The fine-tuning configurations are also taken from the Fairseq repository\footnote{\url{https://github.com/facebookresearch/fairseq/tree/main/examples/roberta/config/finetuning}}: each task uses a batch size of 16 or 32, a learning rate of $1\times 10^{-5}$ or $2\times 10^{-5}$, and is trained for 10 epochs, with the checkpoint achieving the best validation performance selected for testing.

\paragraph{Attention Distillation Setting.} We use the attention weight distillation loss proposed in \cite{hedgehog}: 
\[
\mathcal{L}_{awd}=-\frac{1}{N}\sum_{i=1}^N\sum_{j=1}^N\frac{\exp(\frac{\text{Q}_i^\top \text{K}_j}{\sqrt{d}})}{\sum_{l=1}^N\exp(\frac{\text{Q}_i^\top \text{K}_l}{\sqrt{d}})}\log \frac{\tilde{\phi}_n(\text{Q}_i)^\top \tilde{\phi}_n(\text{K}_j)}{\sum_{l=1}^N\tilde{\phi}_n(\text{Q}_i)^\top \tilde{\phi}_n(\text{K}_l)}.
\]
It computes the cross-entropy between the distribution obtained by normalizing each query with our kernel and the target distribution from softmax attention, then averages this loss over all query positions. During attention weight distillation, we freeze all parameters of RoBERTa, replace its attention computation with our learnable kernel, and train only the kernel parameters for 20,000 steps on a dataset with a batch size of 32 and a learning rate of $1\times 10^{-3}$. For the experiments in \cref{recover}, we perform attention weight distillation followed by fine-tuning directly on the target dataset. In contrast, for \cref{transfer}, we first conduct attention weight distillation on a different dataset and then finetune the model on the target dataset. We select STS-B, QQP, MNLI, and QNLI as distillation datasets because they span multiple domains and are representative of diverse tasks.

\section{Complexity Analysis}
\label{complexity}

\subsection{Time and Space Complexity}
As discussed in \cref{section linear attention}, the overall time and space complexity of linear attention is $\mathcal{O}(Ndd')$ and $\mathcal{O}(Nd+Nd'+dd')$ respectively. 
For Flexformer, $d'=2n$, where $n$ denotes the number of frequencies. Generally, larger $d'$ enhances the expressive power of the learned kernel but increases both parameter count and complexity. A trade-off between expressiveness and efficiency is therefore necessary to determine the optimal hyperparameter $d'$. In subsequent comparative experiments, to ensure maximal fairness, we set $n=d,d'=2d$ following prior work~\cite{RFA,hedgehog}.

\subsection{Additional Parameters} Compared to early linear attention methods~\cite{lineartrans,performer}, more recent studies~\cite{hedgehog,polaformer} incorporate additional learnable parameters to enhance the expressive power of kernel-based linear attention. Flexformer follows the same line. Let $h$ denote the number of heads. Assuming $n=d$, Flexformer requires approximately $d^2$ extra parameters per head for stationary variant ($2d^2$ for nonstationary). To maximize expressiveness, we assume these parameters independent across all heads. In this configuration, Flexformer requires approximately $\frac{1}{12h}$ extra parameters per encoder/decoder layer for stationary variant ($\frac{1}{6h}$ for nonstationary), which accounts for only about $1\%$ of the total parameters when $h = 8$.

\section{Error Bar}
\label{std}
All our main experiments were repeated under three different random seeds, and we report the mean results in the main text. In this section, we additionally report the standard deviations in \cref{lra std}, \cref{wikitext std} and \cref{glue std}.

\begin{table}[htbp]
\centering
\caption{Mean and standard deviation of performance metrics on the LRA benchmarks. All reported metrics represent test accuracy (\%), where higher values indicate better performance. 
The best results for each benchmark are in \colorbox{lightblue}{\textbf{bold}}, 
and the second best in \colorbox{lightgreen}{\underline{underlined}}.}
\label{lra std}
\resizebox{\textwidth}{!}{
\begin{tabular}{lcccccc}
\toprule
Model 
& ListOps & Text & Retrieval & Image & Pathfinder & Average \\
\midrule
Transformer~\cite{transformer} 
& \colorbox{lightgreen}{\underline{37.23$\pm$0.17}} & 65.66$\pm$0.58 & 64.49$\pm$0.61 & 41.89$\pm$0.29 & 73.46$\pm$0.43 & 56.55$\pm$0.21 \\
Reformer~\cite{reformer} 
& 36.62$\pm$0.18 & 63.52$\pm$0.56  & 57.60$\pm$0.58 & 40.06$\pm$0.30 & 73.61$\pm$0.46 & 54.28$\pm$0.22 \\
Longformer~\cite{longformer} 
& 36.46$\pm$0.23 & 63.09$\pm$0.56 & 59.24$\pm$0.60 & 41.23$\pm$0.30 & 72.44$\pm$0.42 & 54.49$\pm$0.21 \\
Linear Trans.~\cite{lineartrans} 
& 17.42$\pm$0.12 & 62.64$\pm$0.54 & 58.92$\pm$0.58 & 43.40$\pm$0.32 & 75.11$\pm$0.44 & 51.49$\pm$0.20 \\
BigBird~\cite{bigbird} 
& 36.88$\pm$0.16 & 65.34$\pm$0.60 & 64.23$\pm$0.60 & 41.93$\pm$0.31 & 73.58$\pm$0.45 & 56.39$\pm$0.23 \\
Performer~\cite{performer} 
& 16.97$\pm$0.14 & 63.67$\pm$0.57 & 55.17$\pm$0.56 & 41.23$\pm$0.29 & 75.38$\pm$0.44 & 50.48$\pm$0.20 \\
RFA~\cite{RFA} 
& 17.20$\pm$0.13 & 64.89$\pm$0.58 & 58.76$\pm$0.61 & 41.00$\pm$0.30 & 74.51$\pm$0.48 & 51.27$\pm$0.21 \\
Skyformer~\cite{Skyformer} 
& 36.71$\pm$0.16 & 65.51$\pm$0.59 & 68.05$\pm$0.63 & 41.36$\pm$0.31 & 74.18$\pm$0.45 & 57.16$\pm$0.23 \\
Cosformer~\cite{cosformer}
& \colorbox{lightblue}{\textbf{37.44$\pm$0.21}} & 65.16$\pm$0.57 & 64.04$\pm$0.61 & \colorbox{lightgreen}{\underline{43.56$\pm$0.30}} & 72.32$\pm$0.42 & 56.50$\pm$0.22 \\
Hedgehog~\cite{hedgehog} 
& 37.02$\pm$0.18 & 64.10$\pm$0.58 & 68.67$\pm$0.63 & 41.64$\pm$0.32 & 75.89$\pm$0.48 & 57.47$\pm$0.23 \\
Polaformer~\cite{polaformer} 
& 36.93$\pm$0.17 & \colorbox{lightblue}{\textbf{69.65$\pm$0.61}} & 65.39$\pm$0.62 & 40.95$\pm$0.29 & 73.26$\pm$0.44 & 57.23$\pm$0.22 \\
\midrule
$\text{Flexformer}_\text{s}$ (Ours) 
& 37.09$\pm$0.18 & \colorbox{lightgreen}{\underline{67.38$\pm$0.60}} & \colorbox{lightgreen}{\underline{71.55$\pm$0.62}} & 43.39$\pm$0.29 & \colorbox{lightgreen}{\underline{76.15$\pm$0.48}} & \colorbox{lightgreen}{\underline{59.11$\pm$0.24}} \\
$\text{Flexformer}_\text{n}$ (Ours) 
& 37.18$\pm$0.18 & 65.72$\pm$0.59 & \colorbox{lightblue}{\textbf{73.94$\pm$0.66}} & \colorbox{lightblue}{\textbf{45.42$\pm$0.32}} & \colorbox{lightblue}{\textbf{77.66$\pm$0.48}} & \colorbox{lightblue}{\textbf{59.99$\pm$0.25}} \\
\bottomrule
\end{tabular}}
\end{table}

\begin{table}[htbp]
\centering
\caption{Mean and standard deviation of PPL on the WikiText-103 development and test sets. 
Lower PPL indicates better performance. 
The best results for each benchmark are in \colorbox{lightblue}{\textbf{bold}}, 
and the second best in \colorbox{lightgreen}{\underline{underlined}}.}
\label{wikitext std}
\resizebox{0.8\linewidth}{!}{
\begin{tabular}{lcccc}
\toprule
 & \multicolumn{2}{c}{\makecell{Small \\ ($41\text{M}$ Params)}} 
 & \multicolumn{2}{c}{\makecell{Big \\ ($247\text{M}$ Params)}} \\
\cmidrule(lr){2-3} \cmidrule(lr){4-5}
Model & Dev. & Test & Dev. & Test \\
\midrule 
Vanilla Trans.          
& \colorbox{lightblue}{\textbf{30.0$\pm$0.06}} 
& \colorbox{lightblue}{\textbf{31.3$\pm$0.07}} 
& 22.6$\pm$0.03 & 23.8$\pm$0.06 \\
\midrule
Linear Trans. & 34.1$\pm$0.12 & 35.6$\pm$0.10 & 27.9$\pm$0.08 & 29.2$\pm$0.08 \\
RFA           & 33.8$\pm$0.11 & 35.3$\pm$0.11 & 26.7$\pm$0.09 & 28.0$\pm$0.10 \\
Cosformer     & 32.8$\pm$0.09 & 34.2$\pm$0.11 & 24.9$\pm$0.05 & 26.2$\pm$0.06 \\
Hedgehog      & 31.7$\pm$0.08 & 33.2$\pm$0.05 & 23.0$\pm$0.06 & 24.2$\pm$0.07 \\
Polaformer    & 31.5$\pm$0.08 & 32.9$\pm$0.08 & 23.1$\pm$0.03 & 24.3$\pm$0.02 \\
\midrule          
$\text{Flexformer}_\text{s}$ 
& 31.1$\pm$0.06 & 32.7$\pm$0.08 
& \colorbox{lightgreen}{\underline{22.2$\pm$0.02}} 
& \colorbox{lightgreen}{\underline{23.4$\pm$0.04}} \\
$\text{Flexformer}_\text{n}$ 
& \colorbox{lightgreen}{\underline{30.3$\pm$0.06}} 
& \colorbox{lightgreen}{\underline{31.6$\pm$0.07}} 
& \colorbox{lightblue}{\textbf{21.2$\pm$0.03}} 
& \colorbox{lightblue}{\textbf{22.3$\pm$0.04}} \\
\bottomrule
\end{tabular}}
\end{table}

\begin{table}[htbp]
\centering
\caption{Mean and standard deviation of performance metrics on GLUE. On CoLA, the Matthews correlation coefficient is reported; 
on STS-B, the average of Pearson and Spearman correlation coefficients is reported; 
for all other tasks, accuracy is reported. Higher values indicate better performance for all metrics. 
The best results of each task are in \colorbox{lightblue}{\textbf{bold}}, 
and the second best in \colorbox{lightgreen}{\underline{underlined}}.}
\label{glue std}
\resizebox{\textwidth}{!}{
\begin{tabular}{lccccccccc}
\toprule
Model & CoLA & SST-2 & MRPC & STS-B & QQP & MNLI & QNLI & RTE & Average \\
\midrule
Performer 
& 22.8$\pm$1.21 & 88.9$\pm$0.32 & 66.3$\pm$0.69 & 38.8$\pm$0.45 & 84.4$\pm$0.40 & 77.5$\pm$0.30 & 76.3$\pm$0.53 & 51.6$\pm$0.26 & 63.3$\pm$0.31 \\
RFA
& 27.6$\pm$0.92 & 90.1$\pm$0.29 & 64.2$\pm$0.71 & 41.2$\pm$0.52 & 86.3$\pm$0.37 & 80.8$\pm$0.29 & 75.8$\pm$0.45 & 53.0$\pm$0.33 & 64.9$\pm$0.29 \\
Hedgehog 
& \colorbox{lightgreen}{\underline{62.5$\pm$0.32}} 
& \colorbox{lightgreen}{\underline{93.3$\pm$0.11}} 
& \colorbox{lightblue}{\textbf{86.0$\pm$0.12}} 
& \colorbox{lightgreen}{\underline{87.7$\pm$0.18}} 
& \colorbox{lightblue}{\textbf{91.2$\pm$0.10}} 
& \colorbox{lightgreen}{\underline{83.7$\pm$0.16}} 
& \colorbox{lightblue}{\textbf{91.2$\pm$0.14}} 
& \colorbox{lightgreen}{\underline{71.0$\pm$0.22}} 
& \colorbox{lightgreen}{\underline{83.3$\pm$0.11}} \\
$\text{Flexformer}_\text{n}$ 
& \colorbox{lightblue}{\textbf{63.3$\pm$0.28}} 
& \colorbox{lightblue}{\textbf{93.6$\pm$0.12}} 
& \colorbox{lightgreen}{\underline{85.8$\pm$0.12}} 
& \colorbox{lightblue}{\textbf{88.3$\pm$0.16}} 
& \colorbox{lightgreen}{\underline{91.1$\pm$0.13}} 
& \colorbox{lightblue}{\textbf{85.1$\pm$0.16}} 
& \colorbox{lightgreen}{\underline{91.0$\pm$0.15}} 
& \colorbox{lightblue}{\textbf{71.5$\pm$0.20}} 
& \colorbox{lightblue}{\textbf{83.7$\pm$0.10}} \\
\bottomrule
\end{tabular}}
\end{table}

\section{Additional Results}
\label{app_d}
In this section, we provide additional experimental results.

\subsection{Hyperparameter Sensitivity}
In all main experiments of this paper, to ensure a fair comparison, we set $n=d,d'=2d$. However, as shown by our analysis in Section 3, a larger $n$ can fit more complex kernel functions, but it also incurs higher computational cost. To guide the practical choice of an appropriate $n$, we investigated how model performance and computational cost vary with $n$ on WikiText-103 and on the Image and ListOps tasks in LRA. 

\begin{figure}[ht]
    \centering
    \includegraphics[width=0.98\textwidth]{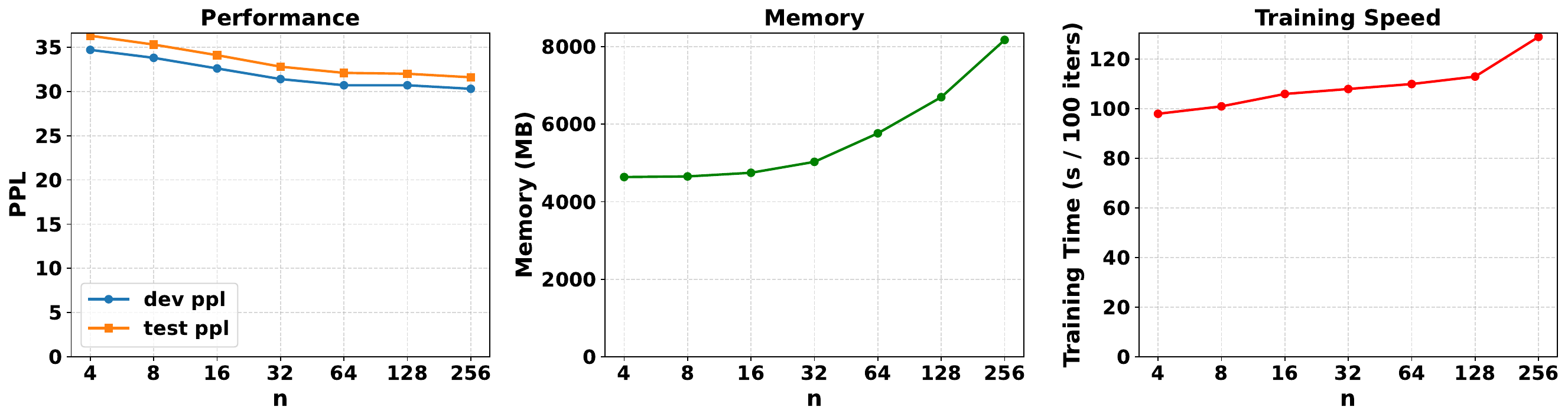}
    \caption{Performance, memory consumption, and training speed of $\text{Flexformer}_{\text{n}}$ in the autoregressive language modeling experiments under different values of the sampling frequency hyperparameter $n$. On large-scale models and datasets, increasing $n$ improves both the fitting and generalization abilities of the model, while also incurring higher computational cost. In the main experiments, we set $n=64$.}
    \label{fig:wiki_n_performance}
\end{figure}

\begin{figure}[ht]
    \centering
    \includegraphics[width=0.98\textwidth]{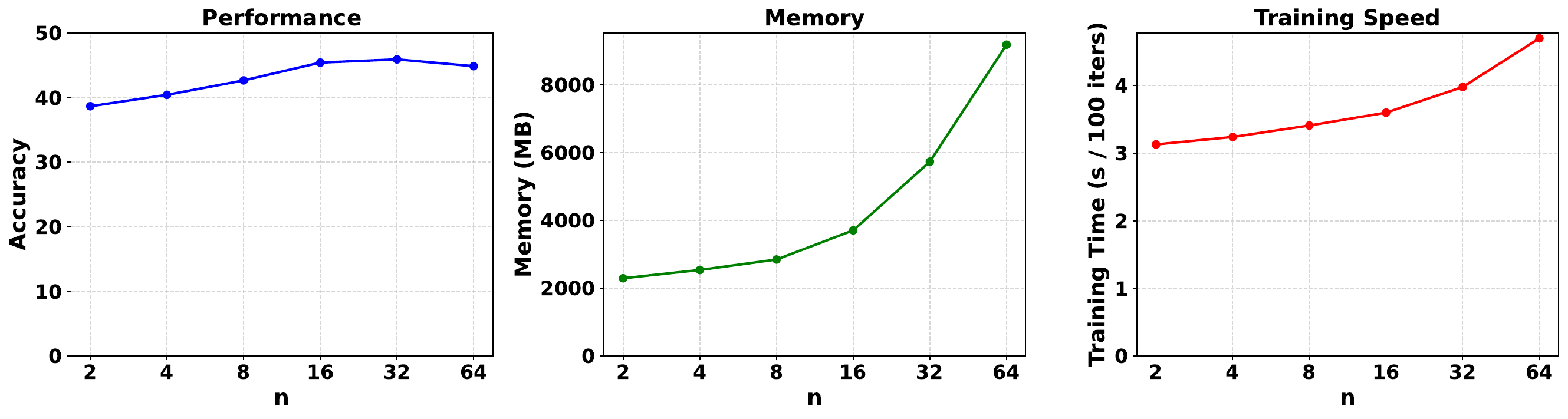}
    \caption{Performance, memory consumption, and training speed of $\text{Flexformer}_{\text{n}}$ on the Image dataset for Long Sequence Classification under different values of the sampling frequency hyperparameter $n$. Both memory usage and training time increase with larger $n$. On the Image dataset, increasing $n$ improves the model’s generalization when $n$ is small; however, as $n$ becomes larger, it increases computational cost and may even lead to overfitting. In the main experiments, we set $n=16$.}
    \label{fig:cifar_n_performance}
\end{figure}

\begin{figure}[ht]
    \centering
    \includegraphics[width=0.98\textwidth]{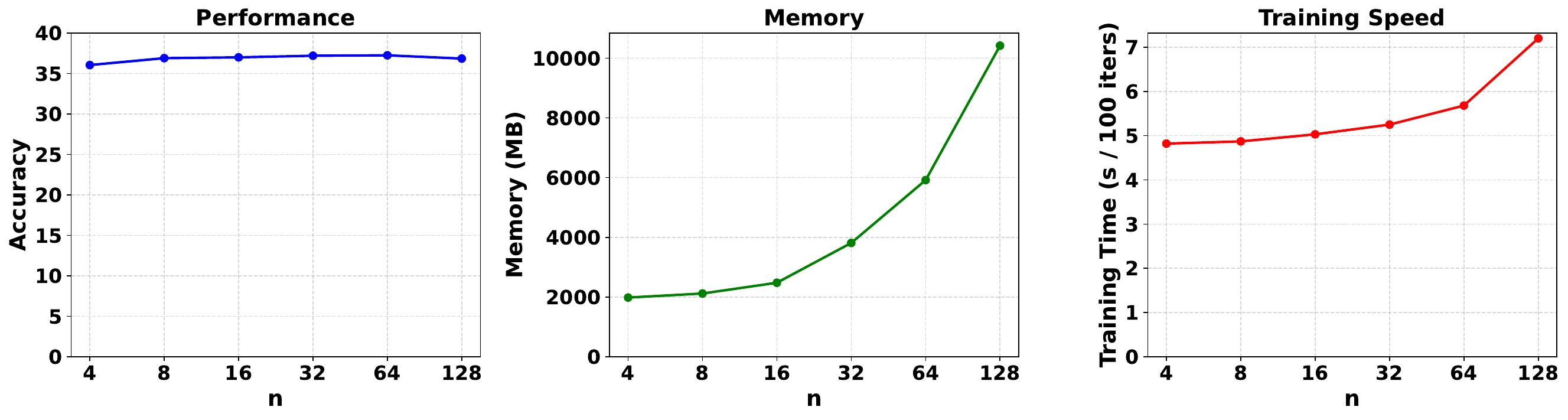}
    \caption{Performance, memory consumption, and training speed of $\text{Flexformer}_{\text{n}}$ on the ListOps dataset for Long Sequence Classification under different values of the sampling frequency hyperparameter $n$. As 
$n$ increases, both memory usage and training time also increase. On ListOps, competitive performance can be achieved even with a very small $n$. In the main experiments, we use $n=64$.}
    \label{fig:listops_n_performance}
\end{figure}

The results in \cref{fig:wiki_n_performance}, \cref{fig:cifar_n_performance} and \cref{fig:listops_n_performance} show that as $n$ increases, the computational cost grows accordingly. On larger-scale models and datasets, increasing $n$ consistently improves generalization, while on smaller-scale benchmarks such as LRA, performance saturates beyond a certain point.  In the main experiments, we set $n=d$ to keep the feature dimension consistent with prior work, under which Flexformer already achieves notable performance gains. Moreover, the additional results show that even with a moderately smaller $n$, Flexformer still outperforms the baselines while maintaining better efficiency.

\subsection{Recovering Softmax Attention}
\begin{figure}[t]
\centering
\begin{subfigure}{0.8\linewidth}
    \centering
    \includegraphics[width=\linewidth]{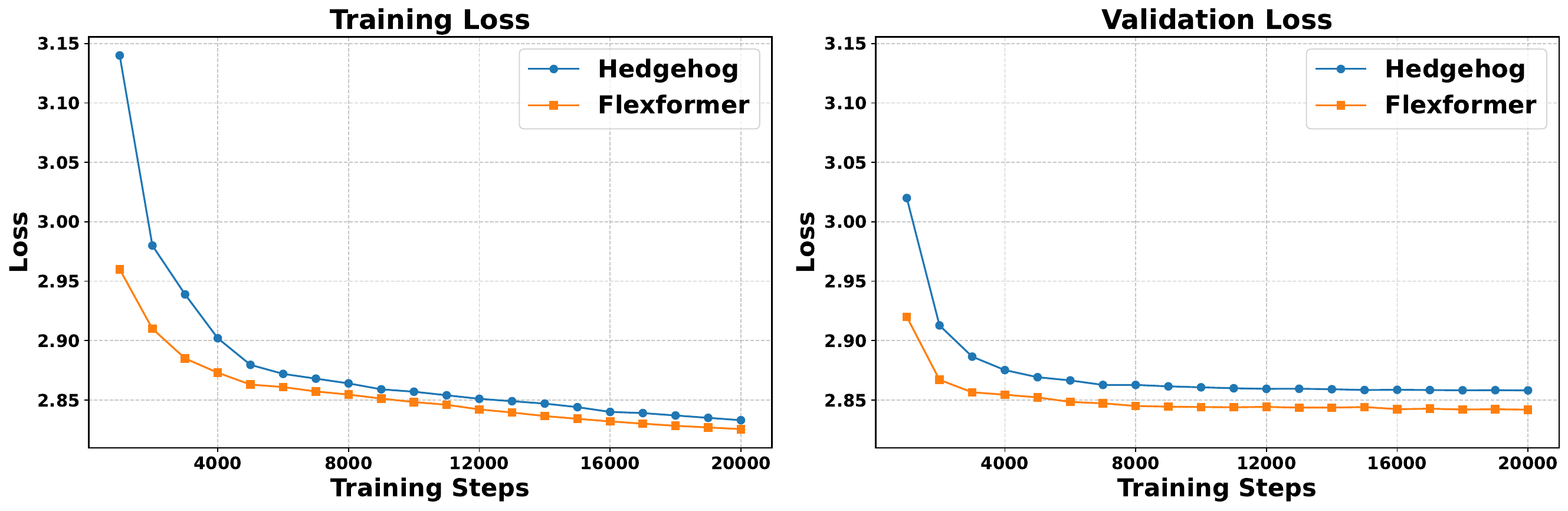} 
    \caption{Distillation on MNLI.}
    \label{fig mnli}
\end{subfigure}

\begin{subfigure}{0.8\linewidth}
    \centering
    \includegraphics[width=\linewidth]{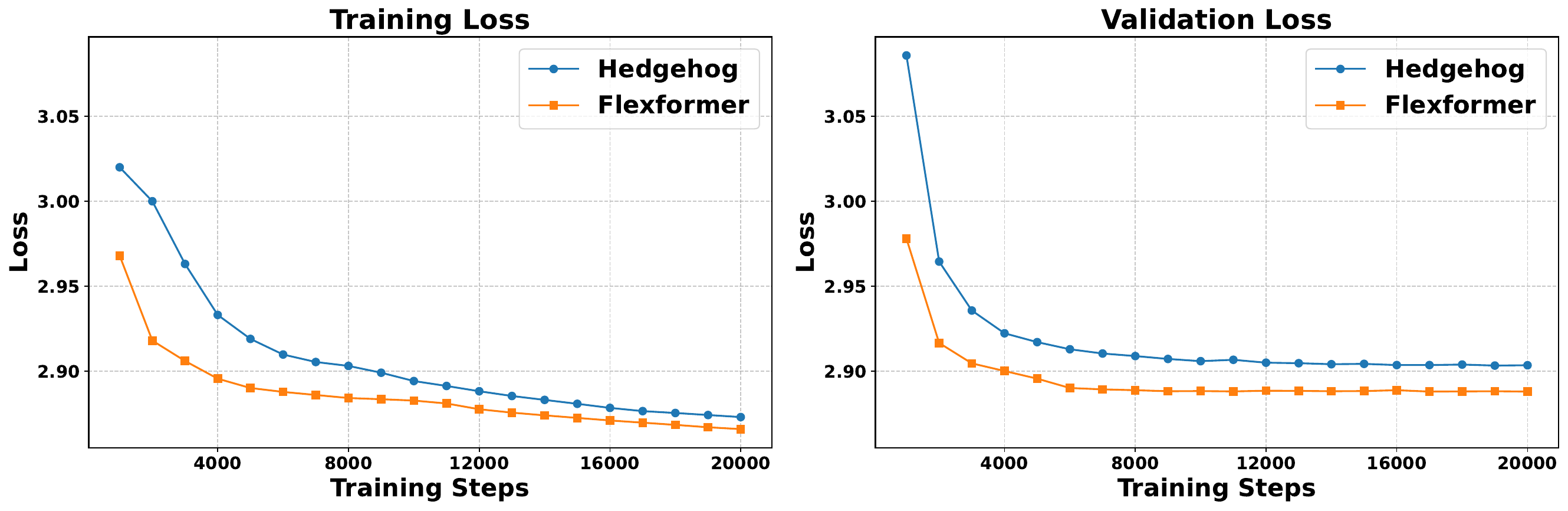} 
    \caption{Distillation on QNLI.}
    \label{fig qnli}
\end{subfigure}

\caption{Comparison of loss curves during attention weight distillation training.}
\label{dist loss}
\end{figure}

In \cref{dist loss}, we selectively compare the loss curves of attention weight distillation using Hedgehog and Flexformer on the MNLI and QNLI datasets. It can be observed that Flexformer consistently achieves lower cross-entropy loss on the validation set compared to Hedgehog. Moreover, Flexformer converges slightly faster than Hedgehog: its validation loss stabilizes around 8,000 training steps, whereas Hedgehog does not stabilize until nearly 12,000 steps. This further demonstrates that the learnable kernel in Flexformer is more flexible. 

\subsection{Transfer Performance}
In \cref{full glue result}, we fully present the results of transferring the model—after performing attention weight distillation on a specific dataset—to other datasets. Flexformer achieves better performance than Hedgehog in almost all cases. Among them, the relative performance improvement reaches up to $13\%$. Moreover, it can also be observed that Hedgehog's performance is relatively unstable when distillation is performed on different datasets.

\begin{table*}[htbp]
\centering
\caption{Comparison of performance when learning the kernel on one dataset and transferring it to other datasets. The best results of each situation are in \colorbox{lightblue}{\textbf{bold}}. }
\label{full glue result}
\small
\begin{tabular}{llcccccccc}
\toprule
Source/Target & Model & CoLA & SST-2 & MRPC & STS-B & QQP & MNLI & QNLI & RTE \\
\midrule
\multirow{2}{*}{STS-B}
 & Hedgehog   & 58.0 & 92.3 & 77.2 & -- & 89.6 & 82.0 & 87.8 & 55.2 \\
 & $\text{Flexformer}_{\text{n}}$ & \colorbox{lightblue}{\textbf{63.5}} & \colorbox{lightblue}{\textbf{93.2}} & \colorbox{lightblue}{\textbf{80.2}} & -- & \colorbox{lightblue}{\textbf{90.7}} & \colorbox{lightblue}{\textbf{84.7}} & \colorbox{lightblue}{\textbf{89.8}} & \colorbox{lightblue}{\textbf{56.7}} \\
\midrule
\multirow{2}{*}{QQP}
 & Hedgehog   & 55.2 & \colorbox{lightblue}{\textbf{93.3}} & 75.2 & \colorbox{lightblue}{\textbf{85.1}} & -- & 83.3 & 87.7 & 50.7 \\
 & $\text{Flexformer}_{\text{n}}$ & \colorbox{lightblue}{\textbf{58.8}} & \colorbox{lightblue}{\textbf{93.3}} & \colorbox{lightblue}{\textbf{79.0}} & 84.8 & -- & \colorbox{lightblue}{\textbf{84.8}} & \colorbox{lightblue}{\textbf{89.1}} & \colorbox{lightblue}{\textbf{53.5}} \\
\midrule
\multirow{2}{*}{MNLI}
 & Hedgehog   & 57.7 & 92.8 & 76.5 & 85.5 & 90.1 & -- & 86.9 & 52.6 \\
 & $\text{Flexformer}_{\text{n}}$ & \colorbox{lightblue}{\textbf{60.6}} & \colorbox{lightblue}{\textbf{93.1}} & \colorbox{lightblue}{\textbf{80.4}} & \colorbox{lightblue}{\textbf{86.7}} & \colorbox{lightblue}{\textbf{90.4}} & -- & \colorbox{lightblue}{\textbf{89.5}} & \colorbox{lightblue}{\textbf{54.9}} \\
\midrule
\multirow{2}{*}{QNLI}
 & Hedgehog   & 51.6 & 91.8 & 76.2 & 85.1 & 89.0 & 82.8 & -- & 53.8 \\
 & $\text{Flexformer}_{\text{n}}$ & \colorbox{lightblue}{\textbf{58.3}} & \colorbox{lightblue}{\textbf{93.2}} & \colorbox{lightblue}{\textbf{78.4}} & \colorbox{lightblue}{\textbf{86.4}} & \colorbox{lightblue}{\textbf{90.6}} & \colorbox{lightblue}{\textbf{84.8}} & -- & \colorbox{lightblue}{\textbf{59.4}} \\
\bottomrule
\end{tabular}
\end{table*}

\section{Limitations and Future Work}
\label{lim_future}

Theoretical analysis in this paper assumes the learned kernel is positive definite, as are the softmax and Gaussian kernels. Empirically, positive definite kernels appear to induce desirable attention behaviors and avoid pathological patterns (e.g., weak self-attention but strong incoming attention). However, whether positive definite kernels are fundamentally optimal for attention mechanisms remains unexplored. We hope future work will provide a deeper understanding of this issue. 

Our experiments are primarily conducted on tasks and datasets related to natural language processing. In fact, there exist numerous other scenarios involving long sequences that require attention mechanisms—for instance, processing high-resolution images or extremely long protein sequences—each of which may demand substantially different attention patterns for effective modeling. Although the LRA benchmark does provide a valuable testbed for evaluating the ability to capture long-range dependencies, thereby complementing the limitation of natural language data (where dependencies are often predominantly local), it remains insufficient to fully reflect the diverse range of attention patterns required across different modalities and domains. We plan to validate the effectiveness of our approach across a broader range of domains in future work.


\end{document}